\documentclass[11pt,a4paper]{article}
\usepackage{times,latexsym}
\usepackage{url}
\usepackage[T1]{fontenc}
\usepackage{times}
\usepackage{latexsym}
\usepackage{xcolor} 
\usepackage{booktabs} 
\usepackage{graphicx}
\usepackage{amsmath}
\usepackage{amssymb}
\usepackage{bbding}

\usepackage[T1]{fontenc}
\usepackage[utf8]{inputenc}
\usepackage{microtype}

\usepackage[acceptedWithA]{tacl2021v1}

\usepackage{xspace,mfirstuc,tabulary}

\newif\iftaclinstructions
\taclinstructionsfalse 
\iftaclinstructions
\renewcommand{\confidential}{}
\renewcommand{\anonsubtext}{(No author info supplied here, for consistency with
TACL-submission anonymization requirements)}
\newcommand{\instr}
\fi

\iftaclpubformat 

\else

\fi

\title{Domain Adaptation in Intent Classification Systems: A Review}

\author{
  Jesse Atuhurra$^\diamond$ Hidetaka Kamigaito$^\diamond$  Taro Watanabe$^\diamond$  Eric Nichols$^\dagger$
\\
  $^\diamond$ NAIST, Japan
  $^\dagger$Honda Research Institute, Japan
  \\
 \texttt{ \{atuhurra.jesse.ag2, kamigaito.h, taro\}@naist.ac.jp   } 
 \\
 \texttt{e.nichols@jp.honda-ri.com}
  \\
}

\date{}

\begin{document}
\maketitle
\begin{abstract}
Dialogue agents, which  perform specific tasks are part of the long term goal of NLP researchers to build intelligent agents that communicate with humans in natural language. Such systems should adapt easily from one domain to another, to assist users to complete tasks. To achieve such systems, researchers have developed a broad range of techniques, objectives, and datasets for intent classification. Despite the progress made to develop intent classification systems (ICS), a systematic review of the progress from a technical perspective is yet to be conducted. In effect, important implementation details of intent classification remain restricted and unclear, making it hard for natural language processing (NLP) researchers to develop new methods. To fill this gap, we review contemporary works in intent classification. Specifically, we conduct a thorough technical review of the datasets, domains, tasks, and methods needed to train the intent classification part of dialogue systems. Our structured analysis describes why intent classification is difficult and studies the limitations to domain adaptation; while presenting opportunities for future work.
\end{abstract}
\begin{figure}[t!]
\includegraphics[width=8cm, height=4.5 cm ]{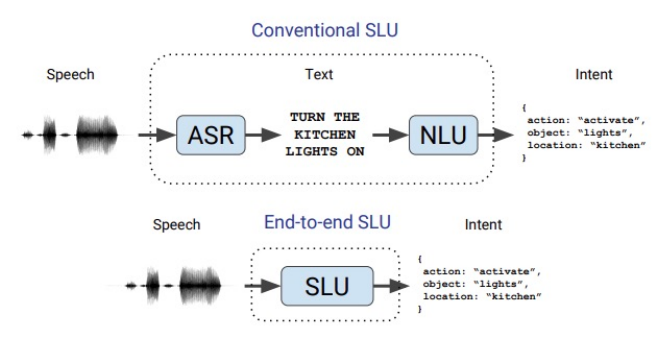}
\centering
\caption{Design choices for a  \emph{spoken language understanding (SLU)} system~\cite{https://doi.org/10.48550/arxiv.1904.03670}. \textit{Top:} Conventional ASR → NLU system for SLU. \textit{Bottom:} End-to-end SLU. At the top, the SLU systems consists of two components: one for \emph{automatic speech recognition (ASR)} and the other for \emph{natural language understanding (NLU)}. The ASR component maps speech data to textual data. Then, the NLU component maps textual data to intent. At the bottom, one trainable model for end-to-end SLU is used to map speech directly to intent.}
\label{fig:tokenized_sentence}
\end{figure}
\section{Introduction}
\label{sec: Introduction}
Task-specific dialogue agents are increasingly widespread because they assist humans to accomplish specific tasks in a particular domain, such as, booking a flight~\cite{5700816}, doing chores inside a home via voicing commands to voice assistants~\cite{https://doi.org/10.48550/arxiv.1805.10190}, among others. These dialogue agents can be spoken to. Hence, dialogue agents can be regarded as spoken language understanding (SLU) systems consisting of two components: automatic speech recognition (ASR), and natural language understanding (NLU). In a conventional SLU system, speech data is mapped to textual data, with the help of an ASR component. Then, textual data is mapped to intent with the help of a NLU component. Alternatively, it is possible to train an end-to-end SLU system where speech data is directly mapped to user-intent. The utterances needed to train the SLU system come from a specific domain, that is, the domain in which the SLU system is likely to be deployed. In short, the job of recognizing user-intent, that is, \textit{intent recognition and classification}, is a central part of dialogue (SLU) agents. 

To achieve effective dialogue agents, the implementation of intent classification involves deploying NLU systems to identify the \textit{intent} of the user. NLU, itself, conveys the machine's ability to understand relevant entities and meaning from textual data. Therefore, the task of intent classification can be defined as; given a set of intent classes (actions), $C$ and a user utterance text, $u$, let the intent classification system (ICS) classify $u$ into one class in $C$, or as out of scope (OOS). Such a task definition for ICS enables the dialogue agent to identify the actions that the user would like to perform. To train intent classifiers, datasets that include utterances (in textual form) containing actions that users perform with that dialogue agent in a specific domain, are required. Moreover, dialogue agents should adapt easily from one domain to another, for them to be more effective. 

Dialogue agents continue to draw attention of NLP researchers, leading to the development of several methods, datasets, and objectives needed to train agents to classify user-intent while performing a task. However, there exists limitations to intent classification when dialogue agents trained in one domain are adapted to another domain. 

To uncover such limitations, first, we conduct a structured analysis of contemporary works on intent classification. In this survey, we explore the datasets, domains, and methods, which have been developed to train the intent classification part of dialogue systems. Moreover, it is clear from our analysis that the advent of Transformers~\cite{https://doi.org/10.48550/arxiv.1706.03762} has led to the development of several transformer-based methods for intent classification. We categorize the contemporary methods for intent classification into; (i) fine-tuning of pretrained language models (PLM), (ii) prompting of PLM, and (iii) few-shot/ zero-shot intent classification. In this survey, we explored all these methods.

Finally, after structured analysis, we point out limitations to domain adaptation. Such limitations manifest in datasets, methods and training objectives used to train dialogue systems for intent classification. The limitations are thoroughly discussed and we present opportunities for future work to overcome the limitations.
Our primary contributions are listed below:
\begin{enumerate}
\item We analyze the datasets used to train intent classifiers and describe the set of domains covered in the datasets. 
\item We systematically study current methods for intent classification and their limitations.
\item We describe future considerations for developing new datasets and methods to overcome the limitations to domain adaptation in intent classifiers.
\end{enumerate}

Our work follows earlier attempts by~\citet{louvan-magnini-2020-recent} to study advances in intent classification and slot filling. However, unlike their work, we present a more recent analysis of intent classifiers; especially the classifiers developed by fine-tuning and prompting of PLM. Such analysis is absent from their study. In addition, we present a detailed review of several NLU datasets on which ICS are trained: with emphasis on the type of data, language representation, and domain coverage for the datasets. Besides, we describe \emph{why intent classification is difficult} and present opportunities for future research to improve intent classifiers.
\begin{table*}[ht!]
\small
\centering
\begin{tabular}{l|l|l|l|l|l}
\toprule
\textbf{Dataset} & \textbf{\#Utterance} & \textbf{\#Domain} & \textbf{\#Intent} & \textbf{\#Slot} &\textbf{\# Lang.}\\
\midrule
ATIS~\cite{5700816}, & 5,871 & 1 & 26 & 129 & 1\\
Almawave-SLU~\cite{Bellomaria2019AlmawaveSLUAN} & 8,542 & 1 & 7 & 39 & 1\\ 
BANKING77~\cite{casanueva-etal-2020-efficient} & 13,242 & 1 & 77 & 83 & 1 \\
CATSLU~\cite{10.1145/3340555.3356098} & 16,258 & 4 &  - & 94 & 1 \\
CLINIC150~\cite{larson-etal-2019-evaluation} & 22,500 & 10 & 150 & \_ & 1\\
Facebook~\cite{https://doi.org/10.48550/arxiv.1810.13327} & 57,049 & 3 & 12 & 11 & 3\\
Fluent Speech Commands~\cite{https://doi.org/10.48550/arxiv.1904.03670} &  30,043 & \_ & 31 & \_ \\ 
HINT3~\cite{arora-etal-2020-hint3} & 1,399 & 3 & 108 & \_ & 1\\
HWU~\cite{https://doi.org/10.48550/arxiv.1903.05566} & 11,106 & 21 & 64 & 54 & 1\\
MASSIVE~\cite{https://doi.org/10.48550/arxiv.2204.08582} & 19,521 & 18 & 60 & 55 & 51\\
Microsoft Dialogue Challenge~\cite{https://doi.org/10.48550/arxiv.1807.11125}  & 38,276  & 3  & 11 & 29 & 1\\
MultiATIS++&  1,422 to & 1 &  21-26 &  99-140 &  9\\
~\cite{https://doi.org/10.48550/arxiv.2004.14353} & 5,897 &&&& \\
MTOP & 15,195 to & 11 & 104-113 & 72-75  & 6\\ 
~\cite{li-etal-2021-mtop} & 22,288 & & & &\\
KUAKE-QIC~\cite{https://doi.org/10.48550/arxiv.2106.08087} & 10,880 & 1 & 11 & \_ & 1\\
SLURP~\cite{slurp} & 16,521 & 18 & 60 & 55 & 1\\
SNIPS~\cite{https://doi.org/10.48550/arxiv.1805.10190} & 14,484 & 1 & 7 & 39 & 1\\
TOP~\cite{gupta-etal-2018-semantic-parsing} & 44,873 & 2 &  25 & 36 & 1\\
Vietnamese JointIDSF~\cite{https://doi.org/10.48550/arxiv.2104.02021} & 5,871  & 1 & 28 & 82 & 1 \\
xSID~\cite{goot2021from} &  10,000 &  7 & 16 & 33 & 13 \\
\hline
**ORCAS-I~\cite{kusa_alexander_de:vries_2022} & 2,000,000 & 126,001 & 1,796,652 & NA & NA\\
**Search4Code~\cite{https://doi.org/10.48550/arxiv.2011.11950} & 1,000,000 & 1 & 8 & NA & NA \\
\bottomrule
\end{tabular}
\caption{\label{Table: datasets_summary}
A summary for NLU datasets. The number of utterances, domains, intents, slots and languages are shown for each dataset. The datasets marked with * are not typical intent classification datasets for a NLU system. \textit{ORCAS-I} is a collection of unique query-urls pairs. \textit{Search4Code} is a dataset built to train models to detect code search intent in search queries for C\# and Java programming languages. \textit{Almawave-slu} is a SLU dataset in Italian. \textit{Vietnamese JointIDSF} is not the real name.  
}
\end{table*}
\section{Task definition: Intent Classification}
\label{sec: Intent_Recognition}
The task of intent classification is formally defined as classifying a user utterance in textual format, $u$, (from a goal-oriented dialog system) into one of the classes $c$ from a set of $N$ pre-defined intent classes (actions), $C$, using an \emph{intent classification} model, $I(u)$.\\
\textbf{Formally:}\\
$\mathbb{C} = \{C_1, C_2, \ldots, C_N \}$, and $I(u) = c$.\\
If \textit{c} is not one of the classes in  $C$, then  \textit{c} is regarded as out of scope (OOS).

The OOS category corresponds to user utterances whose requests are not covered by the intent classification system. Therefore, any utterance is OOS if it does not belong to any of $N$ intent classes, and OOS changes due to $C$. Few-shot and zero-shot intent classification systems aim to predict user utterances in the OOS category.

\paragraph{Example}A pre-trained language model such as BERT~\cite{devlin-etal-2019-bert} was fine-tuned for intent classification. In JointBERT~\cite{https://doi.org/10.48550/arxiv.1902.10909}, BERT was extended to jointly learn intent classification and slot filling. Herein, intent classification predicts the intent label of the query. Slot filling extracts the semantic concepts from the query and attaches the slot label sequence to the sequence of input words. The resulting model is effective at intent classification. Given a sequence of input tokens, $\boldsymbol{x}=(x_1, \ldots, x_T)$, and the output from BERT, $\mathbf{H}=(\boldsymbol{h}_1, \ldots, \boldsymbol{h}_T)$; the hidden state of the first special token $x_1=[CLS]$, denoted $\boldsymbol{h}_1$, was used to predict the intent label, $y^i$, as:
$$p(y^i | \boldsymbol{x}) = softmax_{y^i} (\mathbf{W}^i \boldsymbol{h}_1 + b)$$
The authors fed the final hidden states of other tokens $\boldsymbol{h}_2, \ldots, \boldsymbol{h}_T$ into a softmax layer. They were able to classify over the slot filling labels. For compatibility with WordPiece tokenization, the authors fed each tokenized input word into a WordPiece tokenizer and then used the hidden state corresponding to the first sub-token as input to the softmax classifier.  The slot filling label, $y^s_n$, for a word at the nth position is:
$$p(y^s_n | \boldsymbol{x}) = softmax_{ \boldsymbol{y}^s} (\mathbf{W}^s \boldsymbol{h}_n + b);  n \in 2 \ldots N$$
where $\boldsymbol{h}_n$ is the hidden state corresponding to the first sub-token of word $x_n$, and $N$ is the number of tokenized words in the input.

Finally, they defined a new objective to jointly model intent classification and slot filling:
$$
p(y^i, \boldsymbol{y}^s \mid \boldsymbol{x} )=p(y^i \mid \boldsymbol{x}) \prod_{n=2}^N p(y_n^s \mid \boldsymbol{x} );\\
$$
where $\boldsymbol{y}^s = (y_2^s, ..., y_N^s)$. 

Such a learning objective maximizes the conditional probability $p(y^i,  \boldsymbol{y}^s \mid \boldsymbol{x})$. Hence, the model was fine-tuned end-to-end by minimizing the cross-entropy loss.
\begin{table*}[ht!]
\small
\centering
\begin{tabular}{l|l|l}
\toprule
\textbf{Multilingual Dataset} & \textbf{\# Lang.} &  \textbf{List of Languages}\\
\midrule
Facebook~\cite{https://doi.org/10.48550/arxiv.1810.13327} & 3 & English, Spanish, Thai  \\
\hline
MultiATIS++~\cite{xue-etal-2021-mt5} &  9 & English, Spanish, German, French, Portuguese, \\
& & Chinese, Japanese, Hind, Turkish \\
\hline
MTOP~\cite{li-etal-2021-mtop} & 6 & English, German, French, Spanish, Hindi, Thai\\ 
\hline
xSID~\cite{goot2021from} & 13 & Arabic, Chinese, Danish, Dutch, English, German, \\
& & Indonesian, Italian, Japanese, Kazakh, Serbian, \\
& & Turkish, South Tyrolean \\
\hline
MASSIVE & 51 & Afrikaans, Amharic, Arabic, Azerbaijani, Bengali, Welsh,  \\
~\cite{https://doi.org/10.48550/arxiv.2204.08582} && Danish, German, Greek, English, Spanish, Persian, Finnish,\\
&&   French, Hebrew, Hindi, Hungarian, Armenian, Indonesian, \\
&&  Icelandic, Italian, Japanese, Javanese, Georgian, Khmer,    \\
&&  Kannada, Korean, Latvian, Malayalam, Mongolian, Malay, \\
&&  Norwegian, Dutch, Polish, Portuguese, Romanian, Russian, \\
&&  Slovenian, Albanian, Swedish, Swahili, Tamil, Telugu, Thai,\\
&&  Tagalog, Turkish, Urdu, Vietnamese, \\
&&  Burmese, Mandarin (zh-CN), Mandarin (zh-TW)\\
\bottomrule
\end{tabular}
\caption{\label{Table: datasets_summary_Multilingual}
Multilingual NLU datasets. There is overlap in language representation. English is over-represented. MASSIVE~\cite{https://doi.org/10.48550/arxiv.2204.08582} covers the most languages (that is, 51 languages) by far.}
\end{table*}
\begin{table*}[t!]
\small
\centering
\begin{tabular}{l|l|l}
\toprule
\textbf{Multi-Domain Dataset} & \textbf{\# Domains} &  \textbf{Domain Coverage}\\
\midrule
CATSLU~\cite{10.1145/3340555.3356098} & 4 & map, music, weather, video \\
\hline
CLINIC150 & 10 & banking, work, meta, auto and commute,\\  
~\cite{larson-etal-2019-evaluation} &&  travel, home, utility, kitchen and dining, \\
&& small talk, credit cards \\
\hline
Facebook~\cite{https://doi.org/10.48550/arxiv.1810.13327} & 3 &  weather, alarm, reminder \\
\hline
HINT3& 3 &  mattress products retail, online gaming,\\
~\cite{arora-etal-2020-hint3} && fitness supplements retail \\
\hline
HWU~\cite{https://doi.org/10.48550/arxiv.1903.05566} & 21 &  alarm, audio, audiobook, calendar, \\
&& cooking, datetime, email, game, general, IoT,\\
&& lists, music, news, podcasts, general Q\&A, \\
&& radio, recommendations, social, food takeaway, \\
&& transport, weather \\
\hline
MASSIVE & 18 & * 18 different scenarios recorded from\\
~\cite{https://doi.org/10.48550/arxiv.2204.08582} &&  a home assistant\\
\hline
Microsoft Dialogue Challenge& 3 & movie-ticket booking, taxi booking, \\
~\cite{https://doi.org/10.48550/arxiv.1807.11125} && restaurant reservation \\
\hline
TOP~\cite{gupta-etal-2018-semantic-parsing}, & 11 & alarm, calling, event, messaging, music, news, \\
MTOP~\cite{li-etal-2021-mtop} && people, recipes, reminder, timer, weather\\
\hline
SLURP~\cite{slurp} & 18 & * 18 different scenarios recorded from a home assistant\\
\hline
xSID~\cite{goot2021from} & 7 & music, book, work, restaurant, alarm, reminder, alarm \\
\bottomrule
\end{tabular}
\caption{\label{Table: datasets_summary_Multi_Domain}
Multi-Domain NLU datasets.}
\end{table*}
\section{Datasets}
\label{sec: Datasets}
The datasets are categorized based on data type, multilingualism and domain; and the datasets are shown in Table~\ref{Table: datasets_summary}.
\subsection{NLU and Non-NLU Data}
\label{subsec: NLU_vs_Non_NLU}
The datasets studied in this work are mainly those that contain \emph{textual} or \emph{speech} utterances for training intent classifiers. Amongst datasets shown in Table~\ref{Table: datasets_summary}, two datasets do not contain \emph{textual} or \emph{speech} utterances, despite their use in training intent classifiers. These include: ORCAS-I~\cite{kusa_alexander_de:vries_2022} and Search4Code~\cite{https://doi.org/10.48550/arxiv.2011.11950}. Hence, we designate the datasets as non-NLU data.
\paragraph{ORCAS-I~\cite{kusa_alexander_de:vries_2022}} is a dataset which contains 18 million pairs of unique queries and urls. The ORCAS-I is curated in Italian and this dataset enables the training of a classifier to distinguish between three top level categories of intent, namely: informational, navigational and transactional or to determine the final intent label.\footnote{Details about ORCAS-I can be found here: https://researchdata.tuwien.ac.at/records/pp7xz-n9a06 } 
\paragraph{Search4Code~\cite{https://doi.org/10.48550/arxiv.2011.11950}} is a large-scale web query based dataset of code search queries for C\# and Java. The Search4Code data is mined from Microsoft Bing's anonymized search query logs using weak supervision technique. The main purpose for Search4Code data is to develop NLU models to detect \textit{code search intent} in search queries for C\# and Java programming languages.\\
\subsection{Monolingual and Multilingual Data}
\label{subsec: Multilingual_data}
Almost all the NLU datasets presented in Table~\ref{Table: datasets_summary} contain English utterances. There are few datasets that contain non-English data. \subsubsection{Non-English Monolingual Data}
\label{monolingual_data}
\paragraph{Almawave-SLU~\cite{Bellomaria2019AlmawaveSLUAN}} is designed for training spoken language understanding systems in \textbf{Italian}. Almawave-SLU was derived from the SNIPS~\cite{https://doi.org/10.48550/arxiv.1805.10190} dataset. Almawave-SLU contains 14,484 annotated examples, 7 intents and 39 slots. 
\paragraph{CATSLU~\cite{10.1145/3340555.3356098}} is dataset for training conversational dialogue systems in \textbf{Chinese}. CATSLU contains audio-textual information, spans across multiple domains and domain knowledge. CATSLU is short for Chinese Audio-Textual Spoken Language Understanding. CATSLU contains 16,258 utterances, 83 slots and spans across four domains namely: map, music, weather, and video.
\paragraph{Vietnamese JointIDSF~\cite{https://doi.org/10.48550/arxiv.2104.02021}} is written in \textbf{Vietnamese}. This dataset consists of 5,871 gold annotated utterances with 28 intent labels and 82 slot types which were used to train a \textit{JointIDSF}~\cite{JointIDSF} classifier for intent detection and slot filling in Vietnamese.
\subsubsection{Multilingual Data}
\label{Multilingual_data}
A summary of multilingual NLU datasets is shown in Table~\ref{Table: datasets_summary_Multilingual}. The datasets are discussed below.
\paragraph{Facebook~\cite{https://doi.org/10.48550/arxiv.1810.13327}} contains 57,000 annotated utterances in \textbf{English, Spanish, and Thai.} The number of utterances available for each language is 43000, 8600 and 5000 utterances respectively. The domains covered in this dataset include: weather, alarm, and reminder.
\paragraph{MultiATIS++~\cite{https://doi.org/10.48550/arxiv.2004.14353}} extends the Multilingual ATIS corpus to nine languages across four language families. The nine languages included in MultiATIS++ are: \textbf{English, Spanish, German, French, Portuguese, Chinese, Japanese, Hindi and Turkish.}  For each language in MultiATIS++ (except Hindi and Turkish), there exists 18 intents and 84 slots. There are 5,871 utterances for each language in MultiATIS++, except for Hindi and Turkish which consist of 2,493 and 1,353 utterances respectively.
\paragraph{MTOP~\cite{li-etal-2021-mtop}} consists of 100,000 annotated utterances in six languages namely: \textbf{English, German, French, Spanish, Hindi, and Thai.} MTOP spans across 11 domains and the number of intent types varies between 4 and 27 while the slot types vary between 4 and 17; for various domains. The number of utterances per language in MTOP is: 22,288 for English,  18,788 for German,  16,584 for French,  15,459 for Spanish, 16,131 for Hindi, and  15,195 for Thai.
\paragraph{xSID~\cite{goot2021from}} covers 13 languages from 6 language families. The languages covered in xSID include: \textbf{Arabic, Chinese, Danish, Dutch, English, German, Indonesian, Italian, Japanese, Kazakh, Serbian, Turkish, and South Tyrolean.} xSID includes 10,000 utterances, spans across 7 domains, 16 intents and 33 slots. 
\paragraph{MASSIVE~\cite{https://doi.org/10.48550/arxiv.2204.08582}} is a parallel dataset containing 1,000,000 text utterances that cover 51 languages, span across 18 domains, include 60 intent types, and 55 slots. During the creation of MASSIVE, every utterance was provided in all 51 languages represented in the dataset. For brevity, we refer you to Table~\ref{Table: datasets_summary_Multilingual} for a list of all the languages included in MASSIVE.
\subsection{Domain Coverage}
\label{subsec: Domain_Coverage}
Based on the number of domains covered in each dataset used to train intent classifiers, we categorized the datasets into (i) single domain (ii) multi-domain and (iii) OOS. 
\subsubsection{Single Domain}
\label{subsub: Single_Domain}
As shown in Table~\ref{Table: datasets_summary}, the list of single domain datasets include: ATIS~\cite{5700816}, Almawave-SLU~\cite{Bellomaria2019AlmawaveSLUAN}, BANKING77~\cite{casanueva-etal-2020-efficient}, MultiATIS++~\cite{https://doi.org/10.48550/arxiv.2004.14353}, KUAKE-QIC~\cite{https://doi.org/10.48550/arxiv.2106.08087}, SNIPS~\cite{https://doi.org/10.48550/arxiv.1805.10190}  and Vietnamese JointIDSF~\cite{https://doi.org/10.48550/arxiv.2104.02021}. For example, ATIS is limited to activities related to air travel information, and BANKING77 is limited to banking activities.
\subsubsection{Multi Domain}
\label{subsub: Multi_Domain}
The list of datasets in which the utterances span across multiple domains include: CATSLU~\cite{10.1145/3340555.3356098}, CLINIC150~\cite{larson-etal-2019-evaluation}, Facebook~\cite{https://doi.org/10.48550/arxiv.1810.13327}, HINT3~\cite{arora-etal-2020-hint3}, HWU~\cite{https://doi.org/10.48550/arxiv.1903.05566}, MASSIVE~\cite{https://doi.org/10.48550/arxiv.2204.08582}, Microsoft Dialogue Challenge~\cite{https://doi.org/10.48550/arxiv.1807.11125}, MTOP~\cite{li-etal-2021-mtop}, SNIPS~\cite{https://doi.org/10.48550/arxiv.1805.10190}, SNIPS~\cite{https://doi.org/10.48550/arxiv.1805.10190}, TOP~\cite{gupta-etal-2018-semantic-parsing}, and xSID~\cite{goot2021from}.
A summary for the set of domains covered in the datasets listed above is shown in Table~\ref{Table: datasets_summary_Multi_Domain}.
\subsection{Out-Of-Scope (OOS)}
\label{subsec: Out_Of_Scope_OOS}
\paragraph{BANKING77-OOS~\cite{zhang2022pretrained}} contains both ID-OOS queries and OD-OOS queries in the banking domain. BANKING77 originally includes 77 intents. However, BANKING77-OOS includes 50 in-scope intents in this dataset, and the ID-OOS queries are built up based on 27 held-out semantically similar in-scope intents.
\paragraph{CLINC-Single-Domain-OOS~\cite{zhang2022pretrained}} contains  ID-OOS queries and OD-OOS queries too, in the   banking and credit cards domains.  Each domain in CLINC150 originally includes 15 intents. Each domain in the new dataset includes ten in-scope intents in this dataset, and the ID-OOS queries are built up based on five held-out semantically similar in-scope intents.
\section{Domain Adaptation for Intent Classification}
\label{sec: Intent_Recognition_Methods}
To adapt intent classifiers to new domains, transfer learning, prompting and zero-shot/ few-shot learning have become the standard. Contemporary methods used to classify intent of the user can be broadly categorised into; (i) fine-tuning of PLM, (ii) prompting of PLM, and (iii) few-shot/ zero-shot intent classification. The methods are summarized in Table~\ref{Table: Intent_Classification_Approaches}. \\

\begin{table*}[t!]
\centering
\small 
\begin{tabular}{ p{3cm} | p{12cm}   }
\toprule
Intent Classifiers & Examples of Techniques\\
\midrule
{\bf 1. Fine-tuning} & \\
 \hline
& In-domain to out-of-domain knowledge transfer + Clustering ~\cite{https://doi.org/10.48550/arxiv.2209.06030},\\  
& Joint intent prediction, slot prediction and slot filling ~\cite{https://doi.org/10.48550/arxiv.2207.00828},\\ 
& Transformer compression, Parameter pruning~\cite{https://doi.org/10.48550/arxiv.2203.15610},\\ 
& Plug-and-Play Task-Oriented Dialogue System ~\cite{su-etal-2022-multi}, \\
& Knowledge Distillation~\cite{Jiang2021KnowledgeDF},\\
& Noise-robust intent classification training ~\cite{Sengupta2021}, \\
& Cross-modal intent prediction~\cite{DBLP:conf/interspeech/ChaHJPPK0M21}, \\
& Joint learning phoneme-sequence and ASR transcripts~\cite{https://doi.org/10.48550/arxiv.2102.00804}, \\
BERT & Mahalanobis distance + utterance representations~\cite{https://doi.org/10.48550/arxiv.2101.03778}, \\
& Novel PolicyIE corpus + Seq2seq learning for privacy policies~\cite{ahmad-etal-2021-intent},\\
& New attention layer to use all encoded transformer tokens~\cite{cunha-sergio-etal-2020-attentively}, \\
& Adaptive decision boundary for open intent classification~\cite{https://doi.org/10.48550/arxiv.2012.10209}, \\
& Novel tokens (Observers) + Example-driven training ~\cite{mehri-eric-2021-example}, \\
& Evaluation of Unsupervised pretraining objectives~\cite{Mehri2020DialoGLUEAN},\\
& Intent classification in multi-turn conversations~\cite{senese-etal-2020-mtsi}, \\
& Novel MultiATIS++ corpus + Joint alignment-prediction of slot labels~\cite{https://doi.org/10.48550/arxiv.2004.14353}, \\
& Stacking of MLP + BERT embeddings of incomplete data~\cite{Cunha_Sergio_2021}, \\
& Quantization + Subword-level parameterization~\cite{https://doi.org/10.48550/arxiv.1911.03688}, \\
& Novel benchmark + BERT fine-tuning~\cite{larson-etal-2019-evaluation}, \\
& Joint intent classification and slot filling~\cite{https://doi.org/10.48550/arxiv.1902.10909} \\
\hline
BART & Policy-aware abuse detection~\cite{https://doi.org/10.48550/arxiv.2210.02659}, \\
& mBART fine-tuning on 7 languages ~\cite{FitzGerald2020} \\
\hline
mT5, XLM & Novel MASSIVE corpus + mT5, XLM fine-tuning~\cite{https://doi.org/10.48550/arxiv.2204.08582}, \\
\hline
XLNet & XLNet fine-tuning for citation-impact~\cite{Mercier_2021}, \\
\hline
{\bf 2. Prompting} & \\
\hline
GPT3 & GPT3-prompting + Data augmentation~\cite{sahu-etal-2022-data},\\
\hline
{\bf 3. Zero/Few-shot}  & \\
\hline
& Dual sentence encoders from USE+ConveRT~\cite{casanueva-etal-2020-efficient}, \\ 
& Nearest neighbor classification, Deep self-attention~\cite{zhang-etal-2020-discriminative}, \\
& Self-supervised Contrastive learning,~\cite{DBLP:journals/corr/abs-2109-06349}, \\
& Text augmentation + Self-training~\cite{https://doi.org/10.48550/arxiv.2108.12589}, \\
Few-Shot & Integrate OutFlip-generated OOD samples into training dataset ~\cite{DBLP:journals/corr/abs-2105-05601}, \\ 
& Incremental few-shot learning~\cite{https://doi.org/10.48550/arxiv.2104.11882}, \\
& Data augmentation via neural Example Extrapolation~\cite{DBLP:journals/corr/abs-2102-01335}, \\
& Induction Network + Meta learning~\cite{https://doi.org/10.48550/arxiv.1902.10482}, \\
& Meta-learning + Adaptive metric learning~\cite{yu-etal-2018-diverse}, \\
& Meta-learning for Spoken intent detection~\cite{Mittal_2020}, \\
\hline
& BERT + Adapters~\cite{https://doi.org/10.48550/arxiv.2208.07084}, \\
& Mutual Information Maximization + Prototypical Networks~\cite{Nimah_2021}, \\
& Self-supervised Contrastive learning for dialogue representations~\cite{zhou-etal-2022-learning}, \\
Zero-shot  & Soft Labeling, Manifold Mixup~\cite{9693239}, \\
& Masked LM, Cross-lingual Slot and Intent Detection~\cite{van-der-goot-etal-2021-masked}, \\
& Hierarchical joint modeling for domain and intent classification~\cite{liu2021outofscope}, \\
& Gaussian mixture model, outlier detection algorithm~\cite{yan-etal-2020-unknown},\\
& Capsule networks + Attention~\cite{liu-etal-2019-reconstructing} \\
\bottomrule
\end{tabular}
\caption{\label{Table: Intent_Classification_Approaches} Main approaches to Intent Classification. There are mainly three approaches, namely; fine-tuning of PLMs, prompting of PLMs, and few-shot/zero-shot training. For each approach, examples of techniques used for intent classification are listed. }
\end{table*}
\subsection{Fine-tuning}
\label{Fine_tuning_Intent_Classification}
A number of methods have been developed to identify user intent by fine-tuning the pretrained language models (PLM).
\paragraph{JointBERT~\citet{https://doi.org/10.48550/arxiv.1902.10909}} were the first to propose the fine-tuning of BERT~\cite{https://doi.org/10.48550/arxiv.1810.04805} for intent classification. Prior to that, the attention-based recurrent neural network model~\cite{https://doi.org/10.48550/arxiv.1609.01454} and the slot-gated model~\citep{goo-etal-2018-slot} were the state of the art. The method introduced by~\citep{https://doi.org/10.48550/arxiv.1902.10909} showed that by jointly training the slot filling and intent classification tasks using a PLM, better results are achieved on intent classification accuracy and slot filing F1-score. This was a significant improvement over the best method at the time. 
\paragraph{ConveRT} To reduce the computational cost incurred in training PLM such as BERT,~\citet{henderson-etal-2020-convert} proposed ConveRT. ConveRT employs quantization to reduce the size of transformer network parameters. The effective size of ConveRT after quantization is only 56MB compared to BERT-base (110 MB) and BERT-large (336 MB) based models for intent classification. 
\paragraph{Observers and Example-driven Training} ~\citet{mehri-eric-2021-example} proposed the use of new tokens to replace the [CLS] token in BERT architecture. These new tokens are called Observers. Observers are not attended to and they serve to capture the semantic representation of utterances. Moreover,~\citet{mehri-eric-2021-example} used the Transformer encoder as a sentence similarity model and they compared utterances to examples. This training paradigm, dubbed Example-driven training, enabled the Transformer to learn to classify intent. 
The method was effective in both full data and few-shot settings.  
\paragraph{LightHuBERT} ~\citet{https://doi.org/10.48550/arxiv.2203.15610}  proposed a compression framework for an automatic speech recognition (ASR) BERT-like model, that is HuBERT~\cite{https://doi.org/10.48550/arxiv.2106.07447}. LightHuBERT prunes the structured parameters and employs distillation to retrieve context contained inside HuBERT latent representations. 
\paragraph{DialoGLUE, CONVBERT}~\citet{https://doi.org/10.48550/arxiv.2009.13570} proposed a new benchmark for evaluating natural language understanding tasks, that is, DialoGLUE. They also proposed a BERT-based intent classification model, namely, CONVBERT. 
Performance of the proposed model was evaluated in four settings namely: (1) directly fine-tuning on the target task, (2) pre-training with multilingual language model (MLM) on the target dataset prior to fine-tuning, (3) multi-tasking with MLM on the target dataset during fine-tuning and (4) both pre-training and multitasking with MLM. The performance on zero-shot setting was evaluated too.
\paragraph{Policy-aware Abuse Detection}An intent classification model for Abuse Detection, and based on BART was proposed by~\citet{lewis-etal-2020-bart} to identify abuse in social media posts. 
The proposed method produces human-readable explanations about a post. The classification for abuse is achieved by formulating the policy-aware abuse detection as an intent classification and slot-filing task. Herein, each policy corresponds to an intent which is also associated with a specific set of slots. \citet{https://doi.org/10.48550/arxiv.2210.02659} further created a dataset in which posts were annotated with the slots and then fine-tuned BART to detect abuse.
\paragraph{MASSIVE}~\citet{https://doi.org/10.48550/arxiv.2204.08582} proposed a multilingual intent classification and slot filling model by fine-tuning XLM-Roberta~\cite{conneau-etal-2020-unsupervised} and mT5~\cite{xue-etal-2021-mt5}. Both XLM-R and mT5 were fine-tuned on data available in 51 languages and evaluated, for intent accuracy.
\paragraph{ImpactCite}~\citet{Mercier_2021} proposed ImpactCite for citation impact analysis. To determine the citation intent of a scientific article, they developed an intent classification model based on XLNet~\cite{https://doi.org/10.48550/arxiv.1906.08237}.
\subsection{GPT3-Prompting}
\label{Prompt_based_Intent_Classification}
 The training of PLMs based on prompts has gained traction, and the prompting approach involves the modification of the original input text $x$ using a template to form a textual string prompt that has unfilled slots. Thereafter, the PLM is used to probabilistically fill the unfilled information to obtain a final string, from which the final output $y$ can be derived. 
 
~\citet{sahu-etal-2022-data} achieved data augmentation and generated labelled training data for intent classification by prompting GPT3 to produce utterances for the intents in consideration. 
They showed that GPT3 generates utterances that boost the performance on intent classification for intent classes which are distinct enough. However, the utterances were less effective for intents that are semantically related.
\subsection{Few-shot/ Zero-shot}
\label{Few_Zero_shot_Intent_Classification}
In this section, we describe the latest efforts in few-shot and zero-shot settings to train intent classification models on in-domain out-of-scope (ID-OOS) and out-of-domain out-of-scope (OOD-OOS) data.
\paragraph{Z-BERT-A} To recognize unseen intents in input utterances,~\citet{https://doi.org/10.48550/arxiv.2208.07084} propose Z-BERT-A; an intent classification model based on Transformers~\cite{https://doi.org/10.48550/arxiv.1706.03762}, BERT ~\cite{devlin-etal-2019-bert} and fine-tuned with Adapters~\cite{https://doi.org/10.48550/arxiv.2007.07779}. Z-BERT-A was trained for Natural Language Inference (NLI) and intent classification was cast as an NLI task. 
Then, Z-BERT-A was evaluated for performance on unseen classes. Due to the ability to generate semantically related intents, Z-BERT-A can discover new intents effectively.
\paragraph{DNNC}~\citet{zhang-etal-2020-discriminative} proposed an intent classification model comprising of discriminative nearest neighbor classification with deep self-attention. Given a user-input, they train a binary classifier to identify the training example that matches user utterance. 
This method is based on BERT. The method takes advantage of BERT-style pairwise encoding and better discriminative ability is achieved by transferring knowledge from a NLI model.
\paragraph{CPFT} ~\citet{zhang2021few} proposed a contrastive learning-based method to classify user intent. First, they conducted self-supervised contrastive pretraining, which enabled the model to learn to discriminate semantically similar utterances in absence of any labels. Next, they achieved explicit few-shot intent classification by  performing few-shot intent detection together with supervised contrastive learning. 
This method is effective because contrastive learning pulls utterances from the same intent closer and pushes utterances across different intents farther. 
\paragraph{USE+ConveRT} 
~\citet{casanueva-etal-2020-efficient} proposed a new method that combines USE~\cite{https://doi.org/10.48550/arxiv.1803.11175} and ConveRT~\cite{https://doi.org/10.48550/arxiv.1911.03688} sentence encoders, resulting in dual sentence encoders which are effective for intent classification. 
To detect intent, they used fixed sentence representations encoded by USE and ConveRT. Then, they stacked a Multi-Layer Perceptron (MLP) on top of the fixed representations, followed by a softmax layer for multi-class classification. The resulting model is small in size and effective in few-shot intent classification. 
\section{Why is Intent Classification difficult?}
\label{sec: Intent_classification_difficulties}
In this section, we discuss the challenges faced when developing NLU systems, particularly, the intent classification systems (ICS).
\paragraph{Human-communication is Multimodal} Humans express their \emph{intent} via several communication cues such as; speech, facial expression, eye gaze, gestures, pose, and paralinguistics (that is, tone of voice, loudness, inflection, pitch), among others. As of this writing, the methods necessary to classify user intent based on a combination of these communication cues are yet to be developed. 
\paragraph{Customizability} To achieve an accurate ICS, a high degree of customizability is often required. The uniqueness of each domain results in the development of highly domain-specific ICS. Consequently, such an ICS can not be easily adapted from one domain to another. Moreover, it is common practice to develop an entirely new dataset to train an ICS for a new domain.
\paragraph{Reasoning} To achieve pragmatic language understanding, contemporary ICS ought to depict some \emph{reasoning} ability. Whereas contemporary ICS are built mainly by fine-tuning and prompting of PLMs, it is not yet clear if PLMs exhibit any \emph{reasoning} ability~\cite{https://doi.org/10.48550/arxiv.2108.07258}; or if PLMs simply output co-occurrence patterns in the data on which the PLMs were trained. The PLMs' inability to reason hinders the ICS from reasoning.
\paragraph{Diversity of Natural Language} Humans are gifted with a plethora of languages. It is near impossible to develop ICS, language-agnostic or not, for all the languages. This is due to the expensive cost of computation, datasets and experts required to train ICS in many languages.
\paragraph{Similarity of Intents} As shown by~\citet{sahu-etal-2022-data}, it is more difficult to distinguish between semantically related intents. 
\paragraph{No Training Data} During the initial stages of developing an intent classification  model for a new domain or language, it is possible that there will be little or no training data. This results in very low accuracy on classification of user intent. 
\paragraph{Imbalanced Training Data} The imbalanced nature of the training data in which some intent classes have much more training examples than intent classes results in skewed intent classification performance, in favor of the over-represented intent classes. The creation of a well-balanced training is sometimes difficult because some intents occur much less often than others, resulting in less training examples.
\paragraph{Out-of-Vocabulary} Due to the numerous ways in which users express their intent (for instance, using synonyms), it is inevitable that many words which were unseen by the intent classifier during training will suffice. This would be challenging if the ICS was built on a PLM which was not trained on a large corpus.
\section{Open Issues}
\label{sec: Open_issues}
The open issues that remain due to the difficulties described in~\S\ref{sec: Intent_classification_difficulties} are discussed below. These issues deserve more attention from NLU researchers, for improved intent classification.
\paragraph{Multimodal Input Data} All the ICS covered in this survey are uni-modal, that is, they take in only textual data. Whereas textual input data meets the requirements for ICS to a certain extent, it does not cater for the full range of data modalities that users naturally use to express their intentions when interacting with the ICS. Other modalities include; emotion, pose, speech, gesture. The ability to leverage communication cues as supported by emotion, pose, speech, and gesture would play a significant role in improving the performance of ICS. This calls for the development of ICS that take in multimodal data.
\paragraph{Limitations of Datasets} The list of limitations is shown in Table~\ref{Table: limitations_to_Domain_Adaptation}. 
In summary, the dataset limitations are: (1) unavailability of NLU data in diverse languages, that is, most NLU data is obtainable only in English language; (2) most NLU datasets are designed with a single domain in mind, resulting in single-domain utterances; (3) NLU datasets tend to have a small number of data samples, which are restricted to a small set of specific tasks.
\begin{table}[t!]
\small
\centering
\begin{tabular}{l|l}
\toprule
\textbf{Limitations } & \textbf{Example of Dataset} \\
\midrule
Dataset is monolingual & Almawave-SLU,\\
& CATSLU, \\
& Vietnamese JointIDSF, \\
\hline
Dataset is limited & ATIS,\\ 
to one domain &  Almawave-SLU,\\
& BANKING77, \\
& MultiATIS++,\\ 
& KUAKE-QIC,  \\
& Vietnamese JointIDSF\\
\hline
Few training examples & ATIS\\
\bottomrule
\end{tabular}
\caption{\label{Table: limitations_to_Domain_Adaptation}
A summary of limitations to Domain Adaptation in NLU datasets. Other monolingual datasets (especially those that contain only English data) are shown in Table~\ref{Table: datasets_summary}. }
\end{table}
The restrictive nature of NLU datasets in which a small set of domains is covered makes it difficult to train intent classifiers for new domains.
\paragraph{LM Fine-tuning} First, fine-tuning language models (LM) e.g., BERT is very resource-intensive. Second, in few-shot settings fine-tuning may result in overfitting. Finally, ~\citet{https://doi.org/10.48550/arxiv.1911.03688},~\citet{henderson-etal-2019-training},~\citet{mehri-etal-2019-pretraining} show that general LM-objectives are less effective for intent classification than conversational pretraining based on the response selection task and conversational data.
\paragraph{GPT-Prompting} As shown in ~\citet{sahu-etal-2022-data}, generation of training data using GPT3 is less effective if the intents under consideration are semantically close. 
This calls for fresh ideas on prompting of language models to generate training data in low-data scenarios, for semantically-close classes of intents.
\paragraph{Language Dependence} To a large extent, modern ICS are very specific to one language, usually English. This calls for the development of ICS in other languages, in order for intent classification to be universal to other languages.
\section{Future Direction}
\label{sec: Future_Directions}
To fill the gaps described in~\S\ref{sec: Open_issues}, key factors to consider during creation of new NLU datasets and intent classification methods are discussed below. 
\paragraph{Multimodality} SLU systems typically consist of speech-to-text and text-to-intent pipelines. From this viewpoint, it is reasonable to create new datasets in which both speech data and textual data are well represented. Such multimodal dataset can be used to train a multimodal intent classification system, for both speech and textual inputs. 
\paragraph{Language Diversity} As of this writing, MASSIVE~\cite{https://doi.org/10.48550/arxiv.2204.08582} is the only dataset in which a big number of languages is represented. That is, MASSIVE~\cite{https://doi.org/10.48550/arxiv.2204.08582} is curated in 51 languages. This is only good for a starting point. The languages contained in MASSIVE~\cite{https://doi.org/10.48550/arxiv.2204.08582} mainly come from Europe and Asia, with only one language from Africa (that is, Swahili). 
In comparison, the neural machine translation (NMT) task has several multilingual datasets. For instance, the largest machine translation dataset (that is, CCMatrix) covers 576 language pairs. 
Hence, new datasets that span more languages are yet to be developed for the purpose of training intent classifiers. 
\paragraph{Domain Diversity} From our analysis of NLU datasets presented in this paper, domains covered in NLU datasets include simple in-house activities like setting a reminder, turning-off lights in a house, playing a favorite song, and the like. 
Moreover, HWU~\cite{https://doi.org/10.48550/arxiv.1903.05566} has the highest number of domains represented in the data, that is, 21. These are ideal for voice assistant applications. 
To make intent classification universal to other applications, such as social robots (e.g., a social robot acting as a tutor or life coach), new NLU datasets need to be created. The datasets need to include a diverse range of topics, for instance, related to people hobbies, people personality, emotion utterances, names of places, cuisines, health, occupations, education, and the like. 
\paragraph{Conversational Pretraining} The use of conversational pretraining objectives instead of general LM objectives has immense potential to enable intent classifiers to adapt to new domains, especially in few-shot learning settings. This is because conversational pretraining aligns better with conversational tasks such as dialog act prediction or next utterance generation~\cite{casanueva-etal-2020-efficient}. 
\paragraph{Adapters} For an utterance, the correct intent can be predicted from a set of novel candidate intents by fine-tuning NLI models. To improve performance at inference time,~\citet{https://doi.org/10.48550/arxiv.2208.07084} showed that a BERT-based NLI model can be tuned with adapters. 
They further showed that the NLI model tuned with adapters is effective at zero-shot classification on the candidate intents. 
This approach has a lot of potential because new adapters can easily be added for discovering and classifying intents for unseen intent classes, under new domains.
\paragraph{Contrastive Learning} The recent work by~\citet{zhang2021few} showed that Contrastive Learning has potential to improve few-shot classification of intents. Particularly, contrastive learning brings semantically-related intents closer and pushes unrelated intents further. 
Contrastive learning makes it possible to distinguish between utterances from similar intent and utterances from distant intents. 
To adapt intent classification models to new domains, contrastive learning should be explored further.
\section{Conclusion}
\label{sec: Conclusion}
In this work, we conducted a systematic analysis of intent classification systems. 

By conducting an in-depth study of the datasets and methods used to train intent classifiers, we presented the challenges faced when developing intent classifiers (That is, \emph{Why intent classification or NLU is difficult?}).  The challenges further indicate the limitations to domain adaptation for intent classifiers.

It is eminent that despite the progress made to address such challenges, a few open issues remain. These open issues present NLU researchers with the insights required to develop better intent classifiers, and NLU systems in general. 

We hope our work provides a useful resource to develop new intent classification systems which adapt well to disparate domains.
\newpage
\bibliographystyle{acl_natbib}
\bibliography{main}

\appendix


\end{document}